\documentclass[a4paper]{llncs}

\usepackage{color}
\usepackage[utf8]{inputenc}
\usepackage{amssymb}
\usepackage{graphicx}
\usepackage{tabularx}
\usepackage{textcomp}
\usepackage{listings}
\usepackage{hyperref}
\usepackage{booktabs}
\usepackage{url}
\usepackage{csvsimple}
\usepackage{rotating}
\usepackage[all]{hypcap}  
\usepackage{listings}
\usepackage{amsmath}
\usepackage{multirow}
\usepackage{authblk}
\usepackage{comment}
\usepackage{float}

\newcommand{\keywords}[1]{\par\addvspace\baselineskip
\noindent\keywordname\enspace\ignorespaces#1}

\DeclareGraphicsExtensions{.pdf,.png,.jpg, .jpeg}

\begin{document}

\mainmatter  

\title{Similarity Measure Development for Case-Based Reasoning- A Data-driven Approach}

\titlerunning{Modelling Similarity Measures}

\author{Deepika Verma\inst{1}, Kerstin Bach\inst{1}, and Paul Jarle Mork\inst{2}}

\institute{
Department of Computer Science  \and
Department of Public Health and Nursing 
\\ 
Norwegian University of Science and Technology, Trondheim, Norway \\
\url{http://www.idi.ntnu.no}, \url{http://www.ntnu.no/ism}
}

\maketitle

\begin{abstract}

In this paper, we demonstrate a data-driven methodology for modelling the local similarity measures of various attributes in a dataset. We analyse the spread in the numerical attributes and estimate their distribution using polynomial function to showcase an approach for deriving strong initial value ranges of numerical attributes and use a non-overlapping distribution for categorical attributes such that the entire similarity range [0,1] is utilized. We use an open source dataset for demonstrating modelling and development of the similarity measures and will present a case-based reasoning (CBR) system that can be used to search for the most relevant similar cases.

\end{abstract}

\section{Introduction}

    CBR has gained popularity in the recent years due to its novel approach to abstract and transfer domain-specific expert knowledge into a user-friendly tool which offers appropriate reasoning for solutions to problems ranging from simple daily life tasks to complex tasks which otherwise necessitate expert guidance.
    
      Modelling the local similarities of attributes while preparing a CBR model can be a challenging task for small and simple, and large and complex data sets alike. In this paper, we direct our attention towards the knowledge engineering process of creating a CBR model and present a data-driven approach for modelling local similarity measures using the openly available User Knowledge Modelling dataset\footnote{\url{https://archive.ics.uci.edu/ml/datasets/User+Knowledge+Modeling}} in the myCBR workbench \cite{BachAlthoff2012,myCBR08}. The main contribution of this paper is a methodology for modelling the local similarity measures using a data-driven approach. We will showcase how the knowledge stored in a data set can be leveraged to define strong initial value ranges for both numerical and categorical attributes and therewith moderate and stratify the knowledge modelling process. 
   
    The remainder of this paper is organised into sections as follows: in section \ref{relatedwork}, we discuss related work about the use of data-driven similarity measure development and its application in CBR, followed by section \ref{sim-modelling} wherein we present our similarity modelling approach. Finally, section \ref{conclusion} concludes the work presented in this paper.

\section{Related Work}
\label{relatedwork}

Similar to the preference-based similarity measure development framework presented by authors in \cite{hullermeier_schlegel_2011,abdelhullermeier_2014}, we are presenting a framework for modelling local similarity measures based on the data set available. Therewith we can tailor each similarity measure to the application domain. Using a data-driven approach for automatic similarity learning and feature weighting has been presented by Gabel and Godehardt \cite{gabel_godehardt_2015} where they trained a neural network to induce local and global similarity measures \cite{Richter95}. While we are not automatically assigning the similarity measures, we use the existing cases to derive them.

\section{Data-driven Knowledge Modelling} 
\label{sim-modelling}

In this section, we explain how we implement a CBR system that can be applied to find the most similar and relevant cases. We use the local-global-principle \cite{Richter95} for tailoring the similarity measure for each attribute and thereby build a knowledge model. Once the local similarity measures are defined, we continue to use weighted sum for defining the global similarity.

Some of the most common challenges for utilizing any dataset for developing a CBR system are the identification of suitable dataset context for the problem at hand, definition of initial similarity measures, representation of cases and determination of valuable cases for populating the case base. In this section, we first describe how we populate the case base and generate cases in the developed case representation. Then we present our method for utilizing a given dataset to model the local similarity measures for both numerical as well as categorical attributes.

\subsection{Case Generation }
\label{casegen}

Developing a case representation is the first step of the CBR system development. Depending on the domain and the available data this can be a challenging process on its own. For presenting our data-driven modelling technique, we use the User Knowledge Modelling dataset, which comprises of six attributes, five numerical and one categorical. The description of all the attributes is presented in table \ref{tab:attrdesc}.

\begin{table}[h!]
\centering
\begin{center}
 \begin{tabular}{|p{1,5cm}|p{10cm}|}
 \hline
 Attribute & Description \\ [0.5ex] 
 \hline\hline
 STG & The degree of study time for goal object materials\\ 
 \hline
 SCG & The degree of repetition number of user for goal object materials  \\
 \hline
 STR & The degree of study time of user for related objects with goal object\\
 \hline
 LPR & The exam performance of user for related objects with goal object \\
 \hline
 PEG & {The exam performance of user for goal objects}\\
  \hline
 UNS & The knowledge level of user
 \\[1ex] 
 \hline  
\end{tabular}  
 \caption{Description of attributes in User Knowledge Modelling dataset}
\end{center}
\label{tab:attrdesc}
\end{table}

The categorical attribute \textit{USN} has four permitted values: \textsl{Very Low, Low, Middle, High}. Table \ref{tab:dataset_stats} shows the data statistics of the numerical attributes in the dataset.

\begin{table}[h!]
\centering
\begin{center}
 \begin{tabular}[width=0.8\textwidth]{| c | c | c | c |c |c | } 
 \hline
& STG & SCG & STR & LPR & PEG \\ [1ex] 
 \hline
 count & 403 & 403 & 403 & 403 & 403\\ 
 \hline
 mean & 0.3531 & 0.3559 & 0.4576 &	0.4313 & 0.4563  \\
 \hline
 min & 0 &	0 &	0 &	0 & 0 \\
 \hline
 max & 	0.99 &	0.90 & 0.95 & 0.99 & 0.99\\
[1ex] 
 \hline
\end{tabular}
 \caption{Data set Statistics}
\end{center}
\label{tab:dataset_stats}
\end{table}

\begin{figure}[h!]
    \centering
    \includegraphics[width=0.5\textwidth]{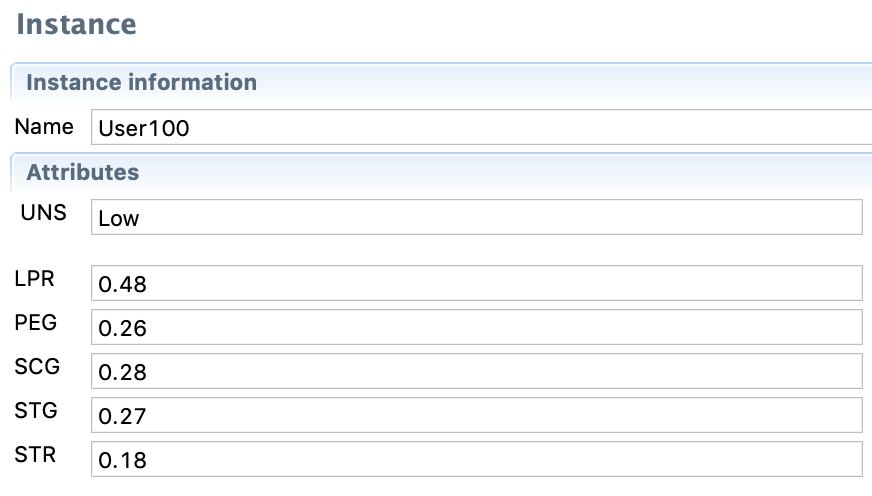}
    \caption{Case representation in myCBR}
    \label{fig:case_rep}
\end{figure}

The case base is then populated by loading the dataset into the previously defined case representation in the myCBR workbench. A single case in myCBR is represented as shown in Figure \ref{fig:case_rep}, where \textsl{User} is the name of the concept which comprises of six attributes present in the original dataset.

\subsection{Data-driven Similarity Measures Development}

The local-global-principle requires both the local similarity measure on the attribute level and the global one on the conceptual to be defined. 

Researchers in CBR domain face the challenge of balancing the input from the domain experts and the available data while modelling the local similarity measures for different attributes in myCBR. Having a criteria which can lead the knowledge modelling process is helpful for both parties. We therefore suggest to make use of the existing data in this process. While setting upper and lower limits for numerical attributes is straight-forward, assigning the similarity behaviour is not. Consecutively, we assume that local similarity measures for continuous numerical attributes are polynomial distance functions (due to their flexibility and better converging ability) and the question is how steep of a similarity decline should be chosen. Therefore, we focus on the polynomial function of the similarity measure for numerical attributes and our goal is to determine their degree. We use box plots for visualizing the distributions and variations in the data set and map this into modelling local similarity measures.

\begin{figure}[h!]
\begin{center}
    \centering
    \includegraphics[width=1\textwidth]{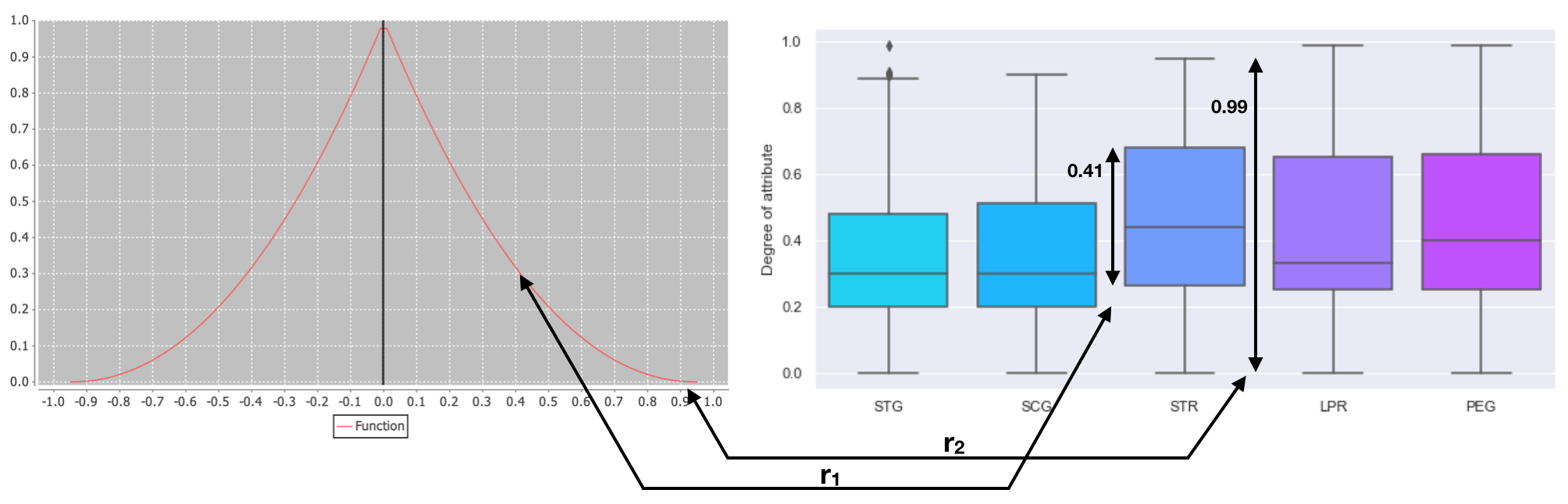}
    \caption{Example for Data-driven Local Similarity Modelling: On the left there is a screen shot of a polynomial similarity function for a value range between 0 and 1. With the arrows we depict how the box-plot for attribute \textit{STR} relates to the decrease in similarity at a certain distance.}
    \label{fig:sims}
\end{center}
\end{figure}

Figure \ref{fig:sims} shows an example of a local similarity measure for a numerical attribute. From there we look into the $Q_1$ and $Q_3$, which indicate the majority spread of the attributes in the data set. In line with \cite{abdelhullermeier_2014,deepika2018}, we decided to take these values as reference points for determining the decrease in similarity.

Hence, creating a box-plot of the data set will allow modelling each attribute since we only take the Inter Quartile Range (IQR) and the range (min to max) into account:

\begin{equation} \label{eq:1}
\begin{split}
 r_1 =  IQR \\
 r_2 = range
 \end{split}
\end{equation}
 
It represents the difference between upper ($Q_3$) and lower ($Q_1$) quartiles in the box-plot, that is $IQR= Q_3 - Q_1$.

We assume that all similarity functions are polynomial and adjust the polynomial degree of the similarity function such that 
\begin{equation} \label{eq:2}
\begin{split}
    y(r_1) \approx 0.30 \\
    y(r_2) \approx 0 
    \end{split}
\end{equation}

We can observe in Figure \ref{fig:sims} how the similarity function varies with respect to the attribute value after applying the methodology in equation \ref{eq:1} and \ref{eq:2}. The bigger the polynomial degree, the steeper the similarity function and more precise the attribute values in retrieved cases. The decline in the similarity function is steeper in the beginning until at $r_1$ it reaches close to $y(r_1)$ and then decreases gradually until at $r_2$ it is approximately close to $y(r_2)$. This way, the similarity function covers the entire attribute range as well as the similarity measure range $[0,1]$. We use this as the initial definition of similarity measures. 

\begin{figure} [!h]
    \centering
    \includegraphics[width=0.6\textwidth]{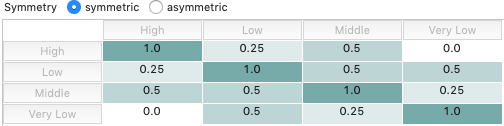}
    \caption{Similarity measure modelling for non-overlapping categorical attribute}
    \label{fig:catsim}
\end{figure}

While the local similarity measures for numerical attributes can be derived using their data distributions, assigning the similarity behaviour for categorical attributes can be challenging as it depends on whether or not there is a pre-existing relationship between the categorical values. In our dataset, the categorical attribute \textit{UNS} has four permitted values which have an implicit relationship amongst each other. The local similarity measure for such an attribute can be modelled such that the relationship amongst the values is preserved while achieving the desired variation in the similarity measure in the range [0,1], as shown in Figure \ref{fig:catsim}. In case of no relationship amongst the values, the similarity of one value to every different value can be set to zero. 

\subsection{Retrieving Similar Cases}
\begin{figure}[h!]
    \centering
    \includegraphics[scale=0.35]{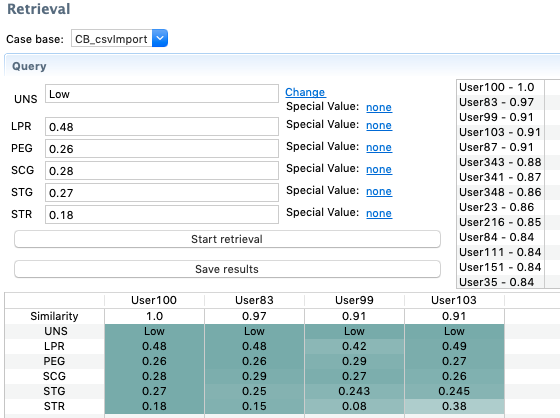}
    \caption{A Query and its retrieval result in the myCBR workbench}
    \label{fig:query}
\end{figure}

Once the casebase and similarity measures are in place, the model can be used to find similar cases. Figure \ref{fig:query} shows the result of one such query retrieval in myCBR. The retrieved cases are sorted by similarity value in descending order, that is, most similar case are displayed at the top while least similar are at the bottom. On the lower part of the figure, the four most similar \textit{Users} are shown in a detailed view. The tool marks closer matches darker.

\section{Discussion and Conclusion }
\label{conclusion} 
In this paper, we have presented an approach to model the local similarity measures of a given dataset in myCBR in a data-driven manner. Our approach can be applied on any dataset to model the similarity measures. A more detailed evaluation of our approach can be found in \cite{deepika2018} where we statistically evaluated its effectiveness using a public health domain dataset and showed that the CBR model created using our approach outperforms the k-NN regressor model in finding the most similar cases.
The approach presented in this work can significantly reduce the efforts required to create new CBR models using different data sets from scratch. Therefore, it is safe to conclude that the approach works well on the used dataset and may also be applicable to other domains.

\bibliography{bibliography}
\bibliographystyle{splncs03}

\end{document}